\PassOptionsToPackage{table}{xcolor}
\PassOptionsToPackage{numbers}{natbib}
\documentclass[sigconf]{acmart}
\AtBeginDocument{%
  }

\copyrightyear{2026}
\acmYear{2026}
\setcopyright{cc}
\setcctype{by-nc-nd}
\acmConference[UMAP '26]{34th ACM Conference on User Modeling, Adaptation and Personalization}{June 08--11, 2026}{Gothenburg, Sweden}
\acmBooktitle{34th ACM Conference on User Modeling, Adaptation and Personalization (UMAP '26), June 08--11, 2026, Gothenburg, Sweden}
\acmDOI{10.1145/3774935.3812717}
\acmISBN{979-8-4007-2311-7/2026/06}

\usepackage[english]{babel}
\usepackage{fixltx2e}
\usepackage{xspace}
\usepackage[table]{xcolor}
\usepackage{graphicx}
\usepackage{booktabs}
\usepackage[numbers]{natbib}
\usepackage{multirow}
\usepackage[english]{babel}
\usepackage{enumitem}
\usepackage{listings}
\definecolor{DarkGreen}{rgb}{0.0,0.5,0.0}
\usepackage{amsmath}
\usepackage{mathrsfs}
\usepackage{caption}
\usepackage{subcaption}
\newcommand{\textcite}[1]{\citeauthor{#1}~(\citeyear{#1})}
\addto\extrasenglish{%
}
\begin{document}

\title[Multi-Dimensional Evaluation of Sustainable City Trips]{Multi-Dimensional Evaluation of Sustainable City Trips with LLM-as-a-Judge and Human-in-the-Loop}

\author{Ashmi Banerjee}
\email{ashmi.banerjee@tum.de}
\affiliation{%
  \institution{Technical University of Munich}
  \city{Munich}
  \country{Germany}
}

\author{Adithi Satish}
\email{adithi.satish@tum.de}
\affiliation{%
  \institution{Technical University of Munich}
  \city{Munich}
  \country{Germany}
}

\author{Wolfgang W\"orndl}
\email{woerndl@in.tum.de}
\affiliation{%
  \institution{Technical University of Munich}
  \city{Munich}
  \country{Germany}
}
\author{Yashar Deldjoo}
\email{yashar.deldjoo@poliba.it}
\affiliation{%
  \institution{Polytechnic University of Bari}
  \city{Bari}
  \country{Italy}
}

\renewcommand{\shortauthors}{Banerjee et al.}
\newcommand{\todo}[1]{\textcolor{red}{TODO: #1}}
\newcommand{\ashmi}[1]{\textcolor{orange}{Ashmi: #1}}
\newcommand{\yashar}[1]{\textcolor{magenta}{YD: #1}}
\newcommand{\adithi}[1]{\textcolor{cyan}{AS: #1}}
\newcommand{\reviewer}[1]{\textcolor{blue}{R: #1}}
\newcommand{\umap}[1]{\textcolor{black}{#1}}
\newcommand{\lbr}[1]{\textcolor{black}{#1}}
\newcommand{\minSign}{\textbf{\textcolor{blue}{$\downarrow$}}}
\newcommand{\maxSign}{\textcolor{blue}{$\uparrow$}}
\newcommand{\maxHighlight}[1]{\textbf{\textcolor{green!70!black}{#1}}}
\newcommand{\minHighlight}[1]{\textbf{\textcolor{blue!70!black}{#1}}}

\newcommand{\gptFour}{\textit{gpt-4o-mini}\xspace}
\newcommand{\gptFive}{\textit{gpt-5}\xspace}
\newcommand{\llama}{\textit{llama-3.2-90b}\xspace}
\newcommand{\geminiOnePointFivePro}{\textit{gemini-1.5-pro}\xspace}
\newcommand{\geminiFlash}{\textit{gemini-2.5-flash}\xspace}
\newcommand{\geminiPro}{\textit{gemini-2.5-pro}\xspace}
\newcommand{\Query}{\ensuremath{\mathbb{Q}}}
\newcommand{\claudeSonnetFour}{\textit{claude-sonnet-4}\xspace}
\newcommand{\deepseek}{\textit{deepseek-v3}\xspace}
\newcommand{\qwen}{\textit{qwen-3-80B}\xspace}
\newcommand{\sigcell}[2]{\cellcolor{green!25}{#1}}
\newcommand{\notsigcell}[2]{\cellcolor{red!25}{#1}} %
\newcommand{\relevance}{\ensuremath{\mathcal{R}}\xspace}
\newcommand{\diversity}{\ensuremath{\mathcal{D}}\xspace}
\newcommand{\popularitymix}{\ensuremath{\mathcal{P}_{\mathrm{M}}}\xspace}
\newcommand{\sustainability}{\ensuremath{\mathcal{S}}\xspace}
\newcommand{\vOne}[1]{\ensuremath{\mathbf{V}_{1}}\xspace}
\newcommand{\vTwo}[1]{\ensuremath{\mathbf{V}_{2}}\xspace}
\newenvironment{promptbox}[1][]{%
  \par\smallskip\noindent
  \begingroup
  \setlength{\fboxsep}{6pt}%
  \fcolorbox{DarkGreen!90}{yellow!10}{%
    \begin{minipage}{\dimexpr\linewidth-2\fboxsep-2\fboxrule\relax}%
    \if\relax\detokenize{#1}\relax\else\textbf{#1}\par\smallskip\fi
}{%
    \end{minipage}%
  }%
  \endgroup
  \par\smallskip
}

\newcommand{\up}{\textbf{\textcolor{green!50!black}{$\uparrow$}}}
\newcommand{\down}{\textbf{\textcolor{red!70!black}{$\downarrow$}}}

\begin{abstract}
Evaluating nuanced conversational travel recommendations is challenging when human annotations are costly and standard metrics ignore stakeholder-centric goals. We study LLMs-as-Judges for sustainable city-trip lists across four dimensions --- relevance, diversity, sustainability, and popularity balance, and propose a three-phase calibration framework: (1) baseline judging with multiple LLMs, (2) expert evaluation to identify systematic misalignment, and (3) dimension-specific calibration via rules and few-shot examples. Across two recommendation settings, we observe model-specific biases and high dimension-level variance, even when judges agree on overall rankings. Calibration clarifies reasoning per dimension but exposes divergent interpretations of sustainability, highlighting the need for transparent, bias-aware LLM evaluation. Prompts and code are released for reproducibility~\footnote{\url{https://github.com/ashmibanerjee/trs-llm-calibration}}.
\end{abstract}

\keywords{LLMs, LLM-as-a-Judge, Calibration, Tourism Recommender Systems, Sustainability}

\maketitle

\section{Introduction and Context}

\lbr{Tourism recommender systems operate in a multi-stakeholder setting, where personalization must balance sustainability, destination capacity, and community impact. Standard accuracy metrics capture relevance but overlook sustainability, overtourism risk, and the balance between popular and lesser-known destinations~\cite{banerjee2023review}.}  
\lbr{Evaluation is further complicated by the tourism data gap: official statistics are often too aggregated for operational recommender systems and differ across countries, limiting cross-country comparability~\cite{lam2013tourism}. No canonical benchmark exists for conversational itinerary recommendation that supports multidimensional evaluation.}  
Expert evaluation is costly and unscalable, while automated metrics often miss contextual and societal factors. LLMs-as-Judges provide scalable approximations of human reasoning but exhibit model-specific biases, especially on subjective criteria~\cite{gu2024survey}.  

\begin{figure}[htbp]
    \centering
    \includegraphics[width=0.75\linewidth,trim={2.5cm 0 2.5cm 0},clip]{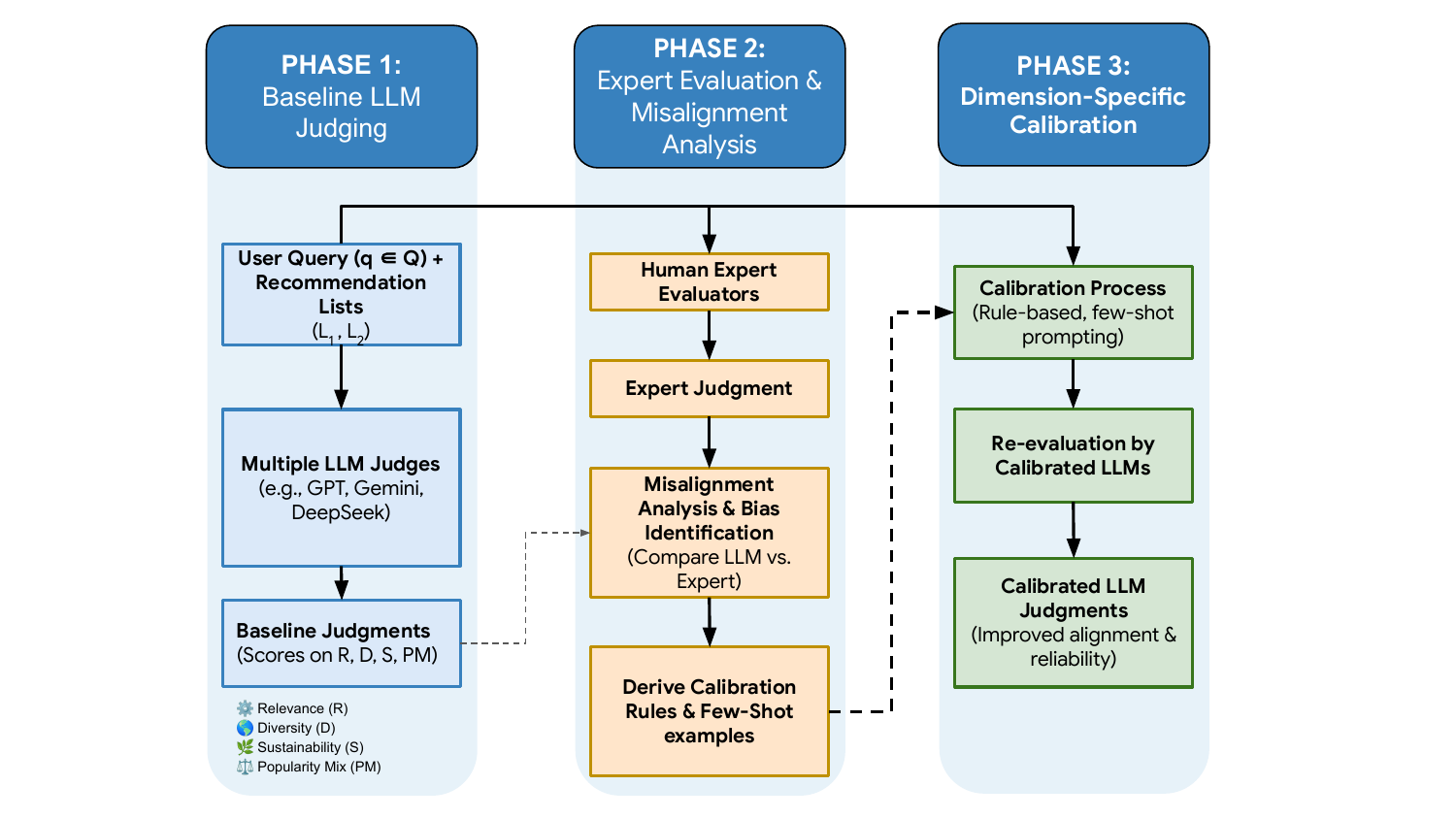}
   \caption{Three-phase LLM calibration framework: baseline judgments, comparison with human evaluations to identify systematic errors, and dimension-specific calibration to improve alignment and reliability.}
    \label{fig:methodology}
\end{figure}
In tourism, this subjectivity is amplified: sustainability varies by seasonality, diversity requires balancing geographic spread with coherence, and popularity must reconcile authenticity with mainstream appeal. Even human evaluators disagree on these dimensions, making perfect consensus neither achievable nor desirable. Moreover, evaluation must consider non-participating stakeholders, such as residents affected by overtourism, aligning with emerging notions of societal fairness~\cite{abdollahpouri2019beyond, banerjee2023review}.

\lbr{To capture these considerations, we evaluate recommendations across four dimensions: \textit{Relevance}, \textit{Diversity}, \textit{Sustainability}, and \textit{Popularity Mix}. Relevance and diversity reflect established yet inherently subjective criteria, shaped by user perceptions and multi-stakeholder trade-offs~\cite{abdollahpouri2019beyond, ekstrand2014user}. Sustainability and popularity mix address the need to balance personalization with destination capacity and community well-being, thereby mitigating overtourism risks~\cite{banerjee2023review}.
In our framework, sustainability is treated as a normative rubric dimension, assessed by expert judgment rather than strict factual verification, with indicators including seasonality, crowding risk, and the encouragement of less commercialized destinations.}
While recent work, such as SynthTRIPs~\cite{banerjee2025synthtrips}, generates diverse, knowledge-grounded travel queries, it does not address how recommendations should be evaluated across these dimensions. This motivates our central question: \emph{How can LLM-based evaluators be calibrated to assess multi-dimensional tourism recommendations in the absence of objective ground truth?}

\textit{Contributions.} We present a methodological study of LLM-based evaluation for sustainable city-trip recommendations:  
\begin{itemize}[noitemsep, topsep=0pt, leftmargin=*]
    \item \textbf{Three-phase calibration framework} aligning LLM judges with human experts across relevance, diversity, sustainability, and popularity balance.
    \item \textbf{Empirical analysis of model biases}, showing calibration improves inter-judge consistency while preserving legitimate disagreement.
    \item \textbf{Lightweight calibration checklist} from human–LLM misalignment to reduce bias without fine-tuning.
    \item \textbf{Reproducible evaluation pipeline} generalizable to other multi-dimensional, domain-specific tasks.
\end{itemize}

\umap{We retain dimension-level evaluation as our primary focus to keep the trade-offs explicit, even though we also compute confidence-weighted aggregated scores. In multi-stakeholder settings, aggregation can obscure trade-offs and mask specific failure modes; analyzing dimensions separately keeps disagreements explicit and actionable.}

\section{Evaluation Framework and Experimental Methodology}
\label{section:framework_experiment}

\umap{%
We propose a three-phase calibration framework to systematically evaluate multi-dimensional European city-trip recommendations using both human experts and LLM judges (\autoref{fig:methodology}). 
In this work, we use the general term \emph{Evaluators} to refer to both LLMs and human experts. When specifying a particular group, we append either \emph{LLM} or \emph{human} explicitly. The term \emph{Judges} is reserved exclusively for LLM evaluators, while \emph{Experts} always refers to human evaluators. 
}

 \begin{figure}[htbp]
    \centering
    \includegraphics[width=0.8\linewidth]{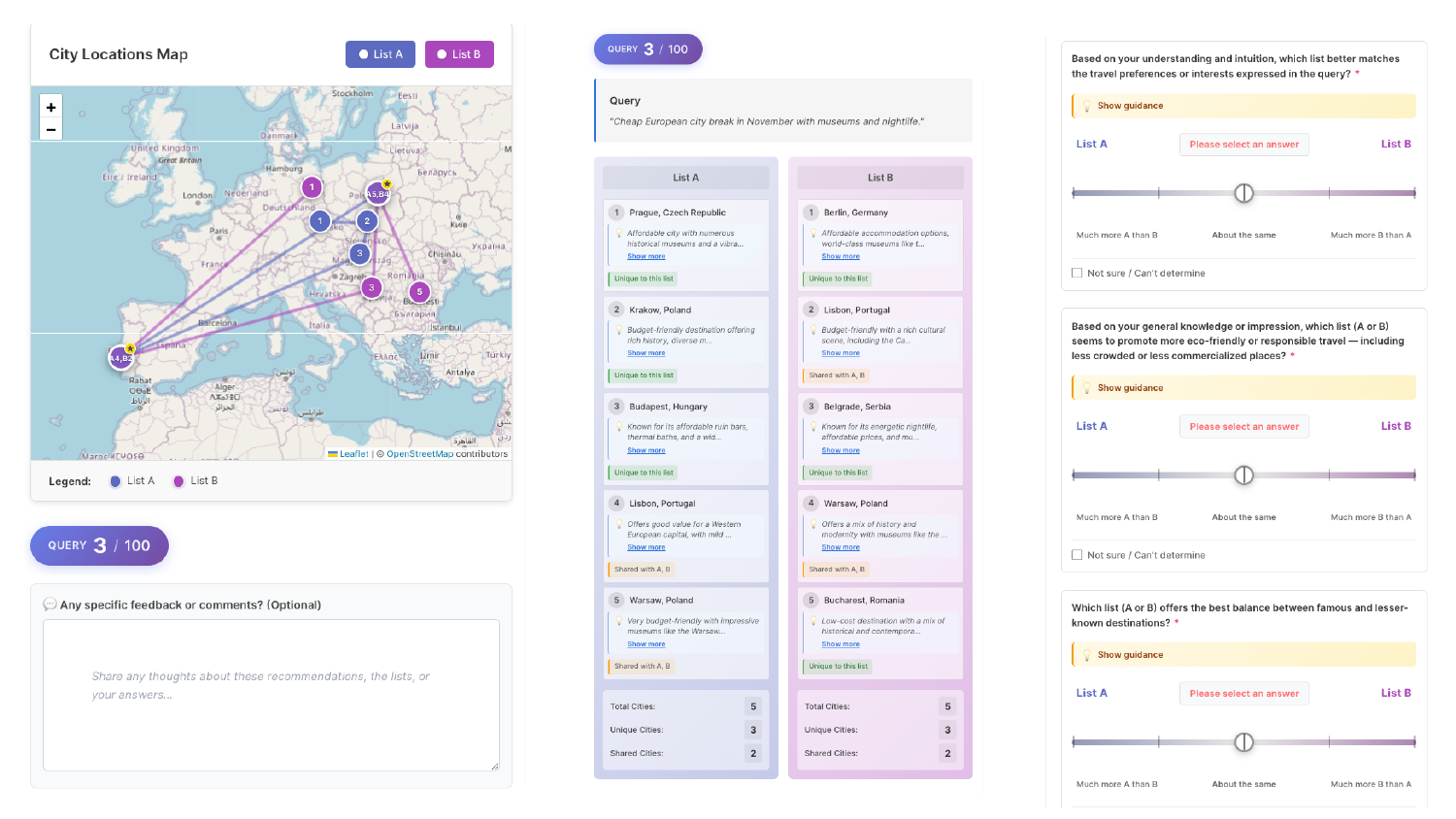}
    \captionsetup{skip=2pt}
    \caption{Web-based survey interface used by human experts for pairwise evaluation of city-trip recommendation lists, including dimension-specific scoring and optional justifications.}
    \label{fig:user_study}
\end{figure}

\lbr{%
Each query $q \in \mathcal{Q}$ represents a traveler request with explicit constraints.
For each query, two LLMs generate ranked lists $L_1$ and $L_2$ of $k=5$ cities, each with justifications.
Evaluators compare these lists using a structured pairwise protocol~\cite{park2024paireval}, assigning dimension-specific judgments over the four predefined dimensions (\autoref{section:eval_dimensions}).
}

\umap{%
\textit{Phase 1: Baseline Judging.}
Paired recommendation lists are independently evaluated by multiple LLM judges, producing baseline judgments across all dimensions.
}

\umap{%
\textit{Phase 2: Expert Evaluation and Misalignment Analysis.}
Baseline judgments are compared with evaluations from five domain experts with at least 5 years of experience in tourism recommendations. Systematic disagreements reveal recurring error patterns and inform calibration rules~\cite{thakur2024judging}.
}

\textit{Phase 3: Dimension-Specific Calibration.}
The derived calibration rules after \emph{Phase 2} and a few-shot examples are integrated into refined judge prompts, and the same queries are re-evaluated to analyze the updated inter-judge variance and alignment with experts. 

\subsection{Evaluation Dimensions} \label{section:eval_dimensions}

\lbr{%
Our study evaluates city-trip recommendation lists generated by LLMs for a natural language query with explicit constraints. }
We evaluate across four dimensions grounded in multi-stakeholder fairness literature~\cite{abdollahpouri2019beyond, ekstrand2014user} and tourism-specific concerns~\cite{banerjee2023review}:
\begin{itemize}[noitemsep, topsep=0pt, leftmargin=*]
    \item \textbf{Relevance (\relevance)} measures how well the recommended cities align with the query's explicit constraints (e.g., budget, season, interests) and implicit preferences inferred from the traveler~\cite{banerjee2025synthtrips,abdollahpouri2019beyond}.
    \item \textbf{Diversity (\diversity)} captures geographic and thematic variety within a list, reflecting the presence of distinct regions and experience types~\cite{abdollahpouri2019beyond,ekstrand2014user}.
    \item \textbf{Sustainability (\sustainability)} evaluates responsible travel characteristics such as seasonality, walkability, public transit access, air quality, and destination capacity. Assessments rely on observable, context-dependent indicators rather than generic or unsubstantiated sustainability claims~\cite{banerjee2023review}.
    \item \textbf{Popularity Mix (\popularitymix)} examines the balance between widely known destinations and lesser-known alternatives~\cite{banerjee2023review}.
    In this paper, \emph{popularity}, \emph{popularity mix}, and \emph{popularity balance} are used interchangeably.
\end{itemize}

\subsection{Experimental Setup}
\label{section:exp_setup}

\paragraph{Query Selection and Data Generation}

Our query set is derived from SynthTRIPS~\cite{banerjee2025synthtrips}, which generates realistic travel requests by combining user preferences with structured constraints (e.g., season, budget, interests, sustainability) and a pool of European cities. We use only the natural-language queries—excluding city lists and constraints—to prompt LLMs (e.g., \textit{“European city break in July, walkable, less touristy”}). From 2,302 generated queries, we retain 123 via deduplication and diversity filtering. Paired recommendations are generated using \gptFour and \geminiFlash; pairs with more than three overlapping cities are discarded, resulting in 100 evaluation queries and one practice query.

\textit{Experimental Conditions.}
We compare \geminiFlash ($L_1$) and \gptFour ($L_2$), and they are evaluated by both Experts and Judges. The LLM judges (\gptFive, \geminiPro, \deepseek) are configured with $\tau = 0.0$, top-$p = 0.95$, and an 8192-token context window. 
Five human experts with at least 5 years of experience in the tourism domain evaluate the recommendations through a web-based interface that displays side-by-side lists with dimension-specific scoring (\autoref{fig:user_study}). Each query receives responses from at least three experts, resulting in approximately 15 hours of total annotation time.
Three LLM judges evaluate the same 100 queries to enable cross-architecture comparison.

\textit{Pairwise Comparison Protocol.}
We use pairwise comparisons to improve judgment stability and reduce scale-related biases~\cite{ekstrand2014user}.  
For each query, evaluators assign dimension-specific scores 
$s_d(L_1, L_2 \mid q) \in \{-2, -1, 0, +1, +2\}$, with positive values favoring $L_1$; responses marked as \emph{unsure} are excluded from the analysis.  
This allows us to assess inter-judge reliability and alignment with human evaluation~\cite{thakur2024judging}, capturing the inherent subjectivity of dimensions such as Sustainability and Diversity.

In addition to dimension-level judgments, LLM judges report a confidence score 
$c_d \in [0,1]$ for each dimension and select a 
\emph{best list} per query to indicate their overall preference.
To quantify confidence-aware overall preference, we compute a 
confidence-weighted aggregated score $S_{\text{agg}}$ for each query:
\begin{equation}
\label{eq:aggregated_score}
S_{\text{agg}} = \sum_{d \in \{\mathcal{R}, \mathcal{D}, \mathcal{S}, \mathcal{P}_M\}} 
s_d \cdot c_d ,
\end{equation}
where $s_d$ denotes the signed pairwise score for dimension $d$ and $c_d$ the 
corresponding confidence. Positive values of $s_d$ favor $L_1$, while negatives favor $L_2$.

At the dataset level, we summarize directional preference using two complementary ratios. The \emph{Aggregated Score Ratio (ASR)} quantifies the total strength of evidence by dividing the sum of all confidence-weighted scores favoring $L_1$ by the sum of those favoring $L_2$. In contrast, the \emph{Best List Ratio (BLR)} measures the frequency of preference, calculated as the total number of times $L_1$ was explicitly selected as the `best' list divided by the total selections for $L_2$.
For both ratios, values $>1$ indicate a net preference for $L_1$, while values $<1$ indicate a net preference for $L_2$.
Together, these metrics allow us to examine whether holistic \emph{best list}
choices align with the direction implied by confidence-weighted, dimension-level
evidence, and to assess consistency between decomposed judgments and overall
preference signals.

\begin{table}[htbp]
\centering
\caption{Calibration rules used to guide LLM judge behavior. The exact prompt formulations are available in the repository\protect\footnotemark.}
\small
\setlength{\tabcolsep}{4pt} %
\renewcommand{\arraystretch}{0.95}
\resizebox{\columnwidth}{!}{%
\begin{tabular}{p{2cm} p{8cm}}
\toprule
\textbf{Rule} & \textbf{Guideline} \\
\midrule
Relevance & Match explicit constraints using city-level evidence; avoid keyword-only matches. \\
Sustainability & Reward verifiable indicators; penalize vague claims; account for seasonality and off-peak travel. \\
Popularity & Favor a balanced mix of mainstream destinations and hidden gems; down-weight crowded options unless justified. \\
Diversity & Ensure geographic and thematic variety unless restricted by the query. \\
Tie-breaking & Resolve close cases using constraint coverage, specificity, factual accuracy, and context. \\
Validation & Verify entity type (city, landmark, country) and factual correctness. \\
Context \& Seasonality & Penalize unsafe, closed, or seasonally unsuitable recommendations. \\
Confidence & Credit evidence-backed claims; mark uncertainty only when ambiguity is unavoidable. \\
\bottomrule
\end{tabular}}

\label{tab:calibration_rules}
\end{table}
\footnotetext{\url{https://github.com/ashmibanerjee/trs-llm-calibration}}

\textit{Calibration Procedure.}
\label{section:calibration_process}
Calibration is performed through iterative re-prompting~\cite{raman2022planning}, informed by recurring patterns of disagreement between LLM judges and human experts. 
It focuses on \emph{systematic} rather than marginal disagreement by analyzing only cases with $|s_d^{\text{llm}} - s_d^{\text{human}}| \geq 2$, which indicate clear directional conflicts with the human majority. Since single-point differences are common across both human-human and judge-judge comparisons and generally reflect acceptable subjective variation rather than errors worth correcting, we only include the two-point disagreements in our misalignment analysis.

We derive eight calibration rules (\autoref{tab:calibration_rules}) that capture operational criteria, common failure modes, and corrective examples which can be generalized across queries and models. 
\lbr{These rules are incorporated into refined judge prompts, along with targeted few-shot demonstrations.}
While the two-point threshold is empirically justified, future work should examine whether stacking multiple calibration rules introduces noise. Overall, our study shows that calibration improves alignment on subjective dimensions and stabilizes inter-judge behavior.

\section{Results and Discussions}

We analyze LLM judge behavior before and after calibration, focusing on alignment with human experts and inter-judge consistency.

\subsection*{RQ1. How well do judges align with human experts, and why do they disagree?}

\textbf{Note on inter-rater reliability.} 
We do not report traditional inter-rater agreement because the number of raters varies across queries and the task involves inherently subjective, multidimensional judgments; instead, we focus on model-specific patterns and calibration effects to analyze consistency and directional preferences.

\textbf{Results.}
Alignment is measured against the majority vote of humans for each response. 
Against the human majority (\autoref{tab:overall_vote}, ``Before''), GPT aligns best on Relevance ($0.53$) and achieves moderate alignment on other dimensions ($0.36$--$0.42$). Gemini shows a comparable macro-average alignment (approximately $0.42$), while DeepSeek performs lower overall (approximately $0.34$), with particularly weak agreement on Relevance (approximately $0.17$).  
Subjective dimensions, especially Diversity and Sustainability, remain the most challenging for judges (mid-$0.30$s to $0.40$s). Human annotators also disagree among themselves, so perfect alignment is neither expected nor always desirable; the objective is to reduce \emph{systematic} LLM errors that humans rarely make.

\begin{table}[h]
\centering
\caption{LLM judge agreement with the human majority vote before and after calibration (values denote the proportion of queries where the judge matches the human majority decision). Arrows indicate the direction of change after calibration.
}
\label{tab:overall_vote}
\begin{tabular}{lcccccc}
\toprule
 & \multicolumn{2}{c}{\textbf{DeepSeek}} & \multicolumn{2}{c}{\textbf{GPT}} & \multicolumn{2}{c}{\textbf{Gemini}} \\
\cmidrule(lr){2-3} \cmidrule(lr){4-5} \cmidrule(lr){6-7}
\textbf{Metric} & Before & After & Before & After & Before & After \\
\midrule
\relevance & 0.17 & 0.18\up & 0.53 & 0.53 & 0.48 & 0.54\up \\
\diversity & 0.35 & 0.29\down & 0.36 & 0.36 & 0.37 & 0.39\up \\
\sustainability & 0.36 & 0.36 & 0.40 & 0.40 & 0.37 & 0.35\down \\
\popularitymix & 0.47 & 0.43\down & 0.42 & 0.42 & 0.45 & 0.39\down \\
\bottomrule
\end{tabular}
\end{table}

\textbf{Why disagreements occur (evidence and examples).}
Manual analysis reveals four recurring causes: 
(i) \emph{Type and factual errors} go unpenalized by judges but are reliably caught by experts: Malta, Santorini, and Plitvice Lakes are treated as cities in some lists. 
(ii) \emph{Context and seasonality}: judges do not flag Lviv during conflict or Keukenhof in October (closed), nor do they penalize Tallinn's deep winter conditions that diminish trip quality; experts do. 
(iii) \emph{Unsure aversion}: Gemini never chooses ``Unsure,'' GPT almost never (one case on Sustainability), while DeepSeek does so more often. 
Forcing a side when the evidence is insufficient reduces alignment on edge cases. 
(iv) \emph{Attention to explicit constraints vs.\ surface cues}: experts reward concrete coverage of budget, time, and persona, while judges sometimes reward keyword echoes or generic sustainability claims.

\vspace{-4pt}

\subsection*{RQ2. What changes after dimension-specific calibration?}
\textbf{Results.}
Calibration primarily stabilizes inter-judge consistency. Gains are largest for Popularity Balance and Sustainability, where concrete rules (e.g., verifying "green" claims) reduce idiosyncratic interpretations. Subjective dimensions like Relevance show mixed results, indicating that rules cannot fully replace world knowledge.
Alignment with human experts improves qualitatively (better reasoning) but quantitative gains are modest (\autoref{tab:overall_vote}). Gemini improves on Relevance (+0.06), while GPT and DeepSeek remain static or dip slightly. This indicates calibration enforces strict, evidence-based assessment—often penalizing vague claims that human experts might forgive—rather than simply mimicking human intuition.
\lbr{We performed statistical tests ($p<0.05$) on pre–post calibration deltas, showing significant improvements primarily in Sustainability for all three judges, with selective gains for DeepSeek on Relevance and Diversity, while Popularity Balance remained stable, indicating that calibration enhances alignment and inter-judge consistency without introducing unintended variance.}

\textbf{Examples.}
Explicit constraints successfully help judges penalize false entities (e.g., Malta as a city) and unsubstantiated sustainability claims. However, trade-offs in Diversity (geographic spread vs. thematic coherence) remain inherently subjective, causing valid disagreements to persist.

\textit{Effects on Directional preferences.}
We analyze whether calibration reduces systematic bias toward specific lists ($L_1$ vs. $L_2$) using confidence-weighted scores (ASR) and best-list selections (BLR) (\autoref{tab:directional_ratios}). The results clarify that calibration reduces uncertainty (noise) but has divergent effects on directionality:
\begin{itemize}[noitemsep, topsep=0pt, leftmargin=*]
    \item \textbf{Correction toward Parity (Gemini)} Pre-calibration, Gemini skewed slightly against $L_1$. Calibration shifts its ASR significantly toward equilibrium ($0.71 \rightarrow 0.98$), aligning with the evidence that the lists are of comparable quality.

    \item \textbf{Preference Sharpening (DeepSeek)} Conversely, DeepSeek amplifies its initial preference for $L_1$ ($1.30 \rightarrow 2.88$). By removing ambiguity, calibration makes the model more decisive in its specific interpretation of the rules (e.g., strictly rewarding specific keywords), leading to polarization ("sharpening") rather than consensus.
    \item \textbf{Entrenchment (GPT)} GPT shows minimal movement ($0.58 \rightarrow 0.64$), suggesting its evaluation heuristics are deeply entrenched and less sensitive to few-shot prompts.
\end{itemize}

\begin{table}[htbp]
\centering
\caption{Directional preference ratios ($>1$ favors $L_1$, $<1$ favors $L_2$) before and after calibration. Arrows indicate the direction of change following calibration, computed using confidence-weighted aggregated scores (ASR) and explicit best-list selections (BLR).}
\begin{tabular}{lcccc}
\toprule
\textbf{Judge} &
\multicolumn{2}{c}{\textbf{ASR}} &
\multicolumn{2}{c}{\textbf{BLR}} \\
\cmidrule(lr){2-3} \cmidrule(lr){4-5}
& Before & After & Before & After \\
\midrule
Gemini   & 0.71 & 0.98\,\up & 0.91 & 1.46\,\up \\
GPT-5    & 0.58 & 0.64\,\up & 0.74 & 0.89\,\up \\
DeepSeek & 1.30 & 2.88\,\up & 1.40 & 2.81\,\up \\
\bottomrule
\end{tabular}

\label{tab:directional_ratios}
\end{table}

\section{Conclusion and Limitations}

Tourism recommendation requires metrics that prioritize sustainability and diversity over simple accuracy. While LLMs enable scalable multi-dimensional assessment, they exhibit model-specific biases. Calibration stabilizes reasoning and harmonizes heuristics, acting as a “clarifier” rather than a universal corrector: it reduces noise while revealing how models interpret rules—moving some toward parity (Gemini) and others toward sharpened preferences (DeepSeek). This transparency shows that one-size-fits-all consensus between judges is unrealistic. Future improvements should refine “Unsure” thresholds and incorporate richer, evidence-linked contextual reasoning. Despite limitations regarding human oversight and normative assumptions, domain-specific calibration provides a scalable, transferable framework, making assumptions explicit and comparable. Our open-source pipeline offers a reusable foundation for transparent, stakeholder-aware evaluation across domains such as health, education, and policy.

\bibliographystyle{ACM-Reference-Format}
\bibliography{main}

\appendix

\end{document}